\documentclass[10pt,twocolumn,letterpaper]{article}

\usepackage{cvpr}              % To produce the CAMERA-READY version

\usepackage{times}
\usepackage{epsfig}
\usepackage{graphicx}
\usepackage{amsmath}
\usepackage{amssymb}
\usepackage{bm}
\usepackage{booktabs}
\usepackage{xspace}
\usepackage{tabularx}
\usepackage{booktabs}
\usepackage{multirow}
\usepackage[ruled,vlined]{algorithm2e}
\usepackage[table]{xcolor}

\definecolor{cvprblue}{rgb}{0.21,0.49,0.74}
\usepackage[pagebackref,breaklinks,colorlinks,allcolors=cvprblue]{hyperref}

\title{Open-Vocabulary Functional 3D Human-Scene Interaction Generation}

\author{
  Jie Liu$^{1,2}$,\; Yu Sun$^{1}$,\; Alp\'{a}r Cseke$^1$,\; Yao Feng$^{4}$,\; Nicolas Heron$^{1}$,\; Michael J. Black$^{3}$, \; Yan Zhang$^{1}$\;   \\
  $^1$Meshcapade,\;
  $^2$University of Amsterdam,\; \\ 
  $^3$Max Planck Institute for Intelligent Systems,\; 
  $^4$Stanford University \\ 
}

\begin{document}

\twocolumn[{%
\renewcommand\twocolumn[1][]{#1}%
\maketitle
\begin{center}
    \centering
    \captionsetup{type=figure}
    \includegraphics[width=\textwidth]{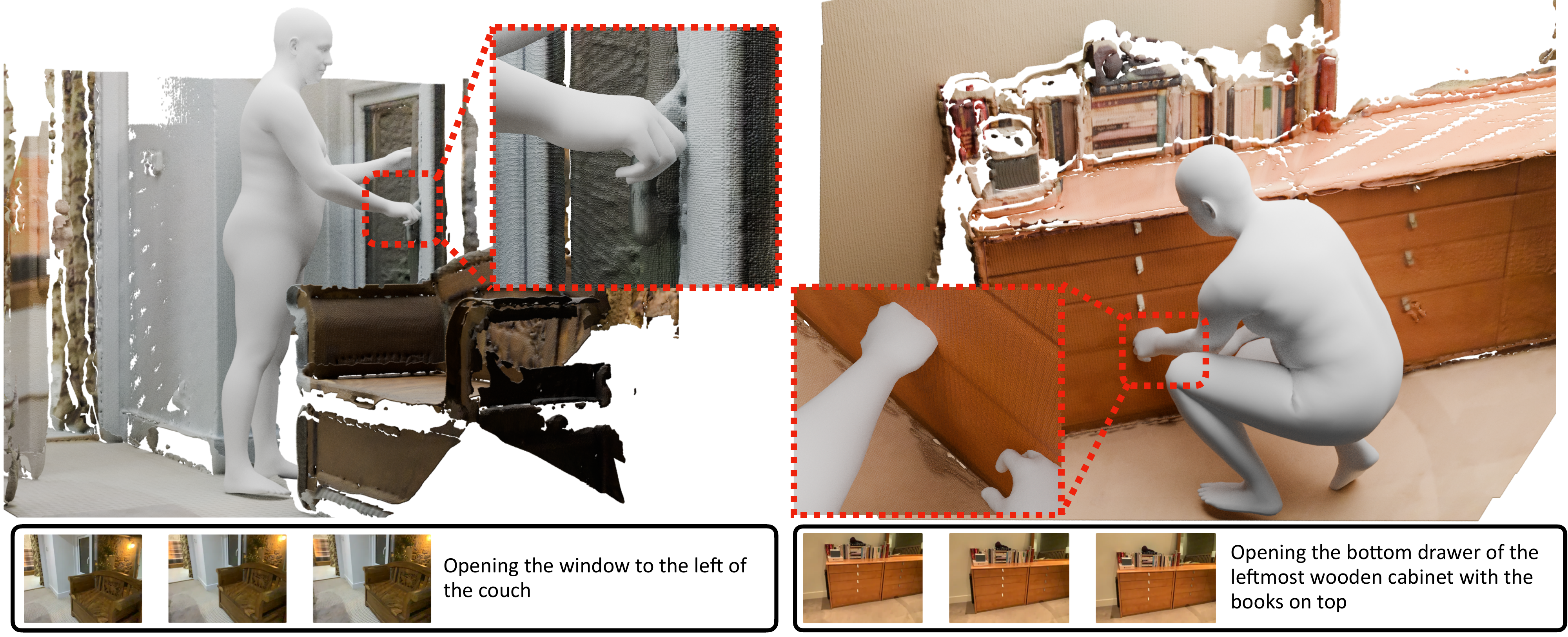}
    \captionof{figure}{Given a set of RGB-D images, their camera parameters, and a task description, our method automatically generates a 3D human body that interacts with the appropriate functional element in the scene. Leveraging the generalization power of foundation models, our method is training-free and applicable to unseen environments and open-vocabulary task descriptions in a zero-shot manner.}
    \label{fig:teaser}
\end{center}%
}]

\maketitle
\begin{abstract}
Generating 3D humans that functionally interact with 3D scenes remains an open problem with applications in embodied AI, robotics, and interactive content creation. 
The key challenge involves reasoning about both the semantics of functional elements in 3D scenes and the 3D human poses required to achieve functionality-aware interaction. 
Unfortunately, existing methods typically lack explicit reasoning over object functionality and the corresponding human-scene contact, resulting in implausible or functionally incorrect interactions.
In this work, we propose FunHSI, a training-free, functionality-driven framework that enables functionally correct human-scene interactions from open-vocabulary task prompts.
Given a task prompt, FunHSI performs functionality-aware contact reasoning to identify functional scene elements, reconstruct their 3D geometry, and model high-level interactions via a contact graph.
We then leverage vision-language models to synthesize a human performing the task in the image and estimate proposed 3D body and hand poses.
Finally, the proposed 3D body configuration is refined via stage-wise optimization to ensure physical plausibility and functional correctness.
In contrast to existing methods, FunHSI not only synthesizes more plausible general 3D interactions, such as ``sitting on a sofa'', while supporting fine-grained functional human-scene interactions, e.g., ``increasing the room temperature''.
Extensive experiments demonstrate that FunHSI consistently generates functionally correct and physically plausible human-scene interactions across diverse indoor and outdoor scenes.
\footnote{Code, models, and more results are available at:\\ \url{https://jliu4ai.github.io/projects/funhsi}}
\end{abstract}

\section{Introduction}
\label{sec:intro}
When asked to ``increase the room temperature'', a human can naturally reason about object functionality, identify the relevant functional element (e.g., a heater knob or thermostat), and interact with it using an appropriate body configuration.
However, performing such functionally-correct interactions in a novel 3D environment remains challenging for embodied intelligent agents, as it requires a holistic understanding of scene semantics and the human actions that the environment affords~\cite{gibson2014ecological, delitzas2024scenefun3d}.
In this work, we investigate to generate realistic and functional interactions between a 3D human body and a novel scene, conditioned on open-vocabulary task descriptions.
An effective solution to this problem  benefits a wide range of applications, including embodied AI, robotics, game production, and video generation, among many others.

The synthesis of 3D human-scene interaction (HSI) has been extensively studied, with existing methods broadly falling into two paradigms.
\emph{Data-driven} approaches learn generative models from paired 3D interaction data, achieving high visual fidelity and realistic human poses in controlled settings.
For example, COINS~\cite{zhao2022compositional} models human body poses conditioned on scene geometry and text commands, while TriDi~\cite{petrov2025tridi} learns a joint distribution over human pose, object geometry, and interaction signals using diffusion models.
Despite their effectiveness, such methods rely on large-scale, high-quality paired interaction datasets and typically require explicit interaction specifications (e.g., ``sitting on a sofa''), limiting their ability to generalize to diverse novel scenes.
To alleviate data dependency, recent work has explored \emph{zero-shot} or \emph{training-free} pipelines that leverage pre-trained vision-language models (VLMs) to generate human-scene interactions.
Representative examples include GenZI~\cite{li2024genzi}, which reconstructs 3D human bodies from multi-view image synthesis, and GenHSI~\cite{li2025genhsi}, which integrates image-based object grounding with 3D body fitting from a single input image.
While these methods improve flexibility and support open-vocabulary task prompts, they are primarily effective for general human-scene interactions describing physical relations or motions, e.g., ``sitting on a sofa'' or ``walking on a bridge''.

In contrast, many real-world tasks like ``open the window'' involve interactions at a functional level, where a human must identify and interact with fine-grained functional elements in the 3D scene to complete the task, such as finding and contacting a window handle to open a window, as shown in Fig.~\ref{fig:teaser}.
We refer to this setting as functional human-scene interaction.
This problem poses fundamental challenges, as it requires reasoning about both the semantics of functional elements in 3D scenes and the 3D human poses needed to establish appropriate contacts.
Existing methods typically lack explicit reasoning about object functionality and the corresponding human-scene contact, leading to interactions that are either geometrically implausible or functionally incorrect.

\begin{figure*}[!t]
	\vspace{-4mm}
	\centering
	\includegraphics[width=\linewidth]{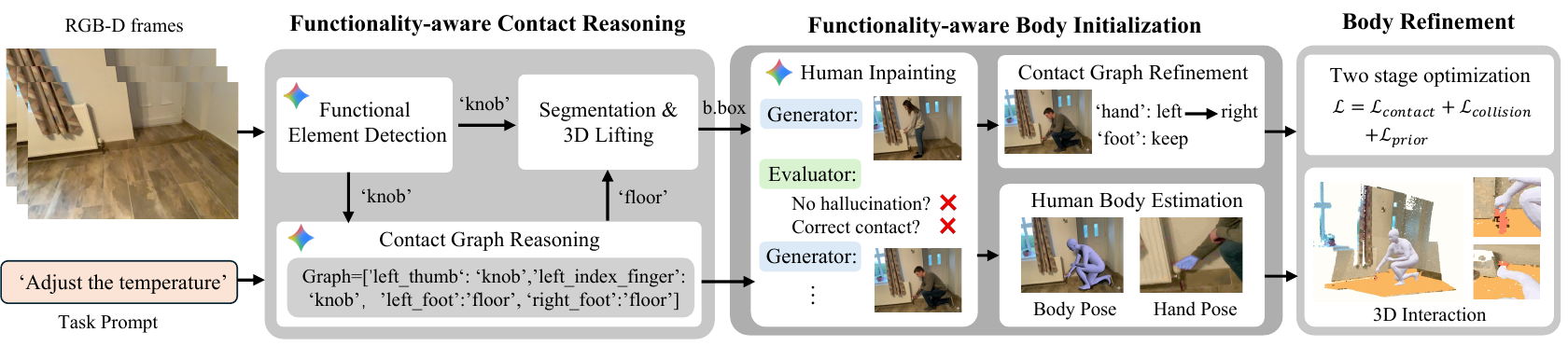}
	\vspace{-6mm}
	\caption{\textbf{Illustration of our FunHSI method.} Given a set of posed RGB-D images, and a task prompt, FunHSI generates 3D humans interacting with functional elements (e.g., ``knob'' or ``switch'') to perform the specified task. First, \textit{functionality-aware contact reasoning} detects elements to be interacted with, constructs a contact graph, and performs segmentation. Next, \textit{functionality-aware body initialization} performs human inpainting, pose estimation, and contact graph refinement, where a generator–evaluator loop ensures no hallucination and correct contact targeting. Finally, \textit{body refinement} performs optimization to improve the body configuration and the contact.}
	\label{fig:framework}
\end{figure*}

In this work, we propose FunHSI, a training-free, functionality-driven
framework that enables functional human-scene interactions from
open-vocabulary task prompts.
Given a set of posed RGB-D images and a task prompt, FunHSI reasons about the functionality of the 3D scene and synthesizes a 3D human that interacts with the scene in a functionally correct manner to accomplish the specified task.
As illustrated in Fig.~\ref{fig:framework}, FunHSI is built upon three key components.
First, we introduce a functionality-aware contact reasoning module to identify task-relevant functional elements in the scene, reconstruct their 3D geometry, and infer high-level interaction patterns via contact graph reasoning.
The resulting contact graph explicitly encodes the contact relationships between the human body and both functional and supporting scene elements, serving as a structured representation that bridges high-level task intent and low-level physical interaction.
Second, we propose a functionality-aware body initialization module that synthesizes a human performing the task in the image and estimates the corresponding initial 3D body and hand poses.
To mitigate hallucinations during human synthesis, we introduce a human inpainting optimization strategy that automatically evaluates and improves the generated human pose configuration.
In addition, since image-based synthesis may produce left-right hand inconsistencies with the inferred contact graph, we further refine the contact graph to align contact specifications with the synthesized human.
Finally, a body refinement module places the initialized 3D human into the scene and performs stage-wise optimization to jointly refine body pose and human-scene contacts, ensuring both physical plausibility and functional correctness.

We conduct experiments on the SceneFun3D dataset~\cite{delitzas2024scenefun3d} under both functional and general human-scene interaction settings.
Extensive qualitative and quantitative results demonstrate the effectiveness of our design and the superior performance of our framework compared to existing baselines.
In addition, we show that FunHSI is compatible with recent feed-forward 3D reconstruction methods, such as MapAnything~\cite{keetha2025mapanything}, and can generate realistic human-scene interactions in reconstructed city scenes.
In summary, our contributions are as follows:
\begin{itemize}[leftmargin=*,topsep=0pt]
\item  We propose FunHSI, a training-free framework that generates functionally correct human-scene interactions from open-vocabulary task prompts. FunHSI extends beyond general interactions to support functional interaction scenarios across diverse scenes and actions.
\item We introduce a robust optimization strategy for inpainting humans and contact graph refinement scheme, providing valuable insights for functional human-scene interactions.
\item Extensive experiments demonstrate that FunHSI achieves strong performance in both functional and general HSI tasks compared to existing baselines. Additionally, FunHSI exhibits strong flexibility and generalization on realistic city scenes captured using smartphones.
\end{itemize}

\section{Related Work}
\label{sec:rw}

\textbf{Data-driven Human-scene Interaction Synthesis}.
Human-scene interaction (HSI) models how humans behave within 3D environments~\cite{zhang2022wanderings, zhao2023synthesizing, yi2024generating, huang2023diffusion, jiang2024scaling, hoifhli}, and many works focus on generating \emph{static} interactions that place the human body into the scene~\cite{savva2016pigraphs, li2019putting, zhang2020generating, hassan2021populating, zhao2022compositional, huang2023diffusion, li2024genzi,li2025genhsi}.
% Representative approaches model body-scene relationships using probabilistic generative frameworks.
A conventional approach is to learn a generative model from paired data.
PLACE~\cite{zhang2020place} employs a conditional variational autoencoder (CVAE) to generate body-scene proximity conditioned on scene geometry, followed by body fitting to produce plausible interactions.
POSA~\cite{hassan2021populating} predicts detailed body-scene contact relations via a graph-based CVAE.
COINS~\cite{zhao2022compositional} incorporates textual prompts to jointly generate pelvis placement and body pose for object-centric interactions.
A closely related research line addresses human-object interaction (HOI), particularly for interactions with small objects where accurate hand-object contact is essential~\cite{taheri2021goal, wu2022saga, li2023object, diller2024cg}.
GOAL~\cite{taheri2021goal} and SAGA~\cite{wu2022saga} first generate target grasping poses and then in-fill motions that reach these targets.
CG-HOI~\cite{diller2024cg} explicitly enforces contact constraints to jointly model human and object motions.
Despite their effectiveness, existing data-driven HSI/HOI approaches rely on large-scale paired interaction data, 
% which is collected using specialized hardware setups, including multi-view RGB-D systems, motion capture, and egocentric devices
~\cite{hassan2019resolving, zhang2022egobody, bhatnagar2022behave, jiang2023full, jiang2024scaling, ma2024nymeria}.
The cost and complexity of acquiring such high-quality multimodal data pose fundamental challenges to scalability and generalization.
% Different from these methods, we leverage the power of pretrained foundation models and propose a training-free method.
% focus on zero-shot human–scene interaction synthesis to overcome the scalability limitations imposed by paired interaction datasets.

\noindent\textbf{Zero-shot HSI Synthesis}
To overcome the data limitation, training-free methods that leverage pre-trained VLMs have been proposed.
GenZI~\cite{li2024genzi} generates 3D bodies based on image generation models. 
Given a description of the task, human pixels are generated individually in tens of images, which are obtained by rendering the same 3D scene from different views. Then the 3D body is reconstructed from the human pixels. 
GenHSI~\cite{li2025genhsi} generates 3D bodies in the scene, which is given by a single image. 
Given the text description, the object to be interacted with is segmented in the image and is lifted to a 3D mesh. 
InterDreamer~\cite{xu2024interdreamer} performs high-level planning to translate a freeform task description into text descriptions of existing text-to-motion datasets.
ZeroHSI~\cite{li2024zerohsi} first combines a body~\cite{li2024animatable}, an object, and a scene together, and renders an image via Gaussian spatting as the first HSI frame. Then video generation produces future frames, from which the camera, object and body motions are estimated.
Despite their progress, existing methods often fail to produce \emph{functional} human-scene interactions with both body-scene and detailed hand-object interactions.
In contrast, our method understands the object functionality and produces functional HSIs. 
For example, given the prompt ``open the door,'' our method  automatically identifies the doorknob and synthesizes a 3D human manipulating the doorknob.

\noindent\textbf{Functional 3D Scene Understanding}
3D scene understanding aims to assign semantic labels to scene elements~\cite{schult2022mask3d, zhong2022maskgroup, graham20183d}.
To support complex reasoning on 3D scenes, large language models (LLMs) have been fine-tuned with language-scene paired data~\cite{deng20253d, zhi2025lscenellm, mei2025perla, Zhu_2025_ICCV, kang2025robin3d}.
However, 3D LLMs remain less mature than 2D VLMs due to data scarcity and computational cost.
To better exploit the power of 2D foundation models, several approaches perform reasoning in posed RGB-D images and then lift the results into 3D space.
OpenScene~\cite{peng2023openscene} back-projects dense 2D features into 3D using known camera parameters, enabling zero-shot open-vocabulary object and affordance grounding in point clouds.
OpenMask3D~\cite{takmaz2023openmask3d} also uses this paradigm for open-vocabulary 3D instance segmentation.
% by jointly leveraging point clouds and posed RGB-D images.
Beyond semantic segmentation, recent works investigate \emph{functionality understanding}, which models how objects or regions can be interacted with or used~\cite{delitzas2024scenefun3d, corsetti2025functionality, zhang2025open}.
SceneFun3D~\cite{delitzas2024scenefun3d} introduces functionality segmentation and curates a multimodal dataset with high-fidelity point clouds, RGB-D images, and language task annotations.
Fun3DU~\cite{corsetti2025functionality} proposes a training-free approach for functionality segmentation using LLMs.
FunGraph3D~\cite{zhang2025open} predicts functional 3D scene graphs by detecting interactive elements and inferring their relationships.
In this work, we not only perform functional scene understanding but also synthesize a 3D human performing the relevant task.

\section{FunHSI} 
\label{sec:method2}
As shown in Fig.~\ref{fig:framework}, FunHSI takes as input a set of posed RGB-D images and a task prompt, and generates a 3D human performing task-specific interactions with the scene.
Overall, FunHSI consists of three key modules.
First, the \textit{functionality-aware contact reasoning} module (Sec.~\ref{sec:4.1}) identifies task-relevant functional elements in the scene, reconstructs their 3D geometry, and performs contact graph reasoning to produce the high-level interactions.
Second, the \textit{functionality-aware body initialization} module (Sec.~\ref{sec:4.2}) leverages the inferred functional elements and contact relations to synthesize a human in the image and estimate the 3D body and the hand poses.
Finally, the \textit{body refinement} module (Sec.~\ref{sec:4.3}) places the initialized 3D body into the 3D scene and performs stage-wise optimization to refine the body and hand poses, and human-scene contacts.

\subsection{Preliminaries}
We denote the SMPL-X model~\cite{SMPL-X:2019} as $\mathcal{M}(\beta, r, \varphi, \theta)$, 
where $\beta\in\mathbb{R}^{10}$ denotes the shape parameters,
$r\in\mathbb{R}^3$ the root translation,
$\varphi\in\mathbb{R}^3$ the root orientation,
and $\theta=[\theta^{b},\theta^{h}]$ the pose parameters.
Here, $\theta^{b}\in\mathbb{R}^{63}$ and $\theta^{h}\in\mathbb{R}^{90}$ represent the body and the hand poses, respectively.
% All rotations are represented in axis-angle form.
Given these body parameters, it can produce a body mesh with 10,475 vertices via forward kinematics (FK).
In addition, the body signed distance field (SDF), denoted as $\Psi(\beta, r, \varphi, \theta)$, is computed via VolumetricSMPL~\cite{mihajlovic2025volumetricsmpl}.
Since both $\mathcal{M}(\cdot)$ and $\Psi(\cdot)$ are differentiable, provided external constraints on the body, inverse kinematics (IK) can be performed via backpropagation to optimize the body parameters and the contacts.

\subsection{Functionality-aware Contact Reasoning}
\label{sec:4.1}
Since the task prompt typically specifies a high-level goal without explicitly describing which elements to interact with or how the interaction should be carried out, FunHSI must automatically reason about scene functionality, identify task-relevant functional elements, and infer appropriate contact relations with the human body.
Accordingly, this module consists of two stages: \textit{functionality grounding and reconstruction} and \textit{LLM-based contact graph reasoning}.

\paragraph{Functionality grounding and reconstruction.}
Given a task prompt such as ``adjust the temperature'', we first identify task-relevant functional elements in the RGB images using a vision-language model (VLM).
In our implementation, we employ Gemini-2.5-Flash~\cite{comanici2025gemini} to infer candidate functional elements conditioned on the task description.
Based on the task prompt and the inferred functional elements, 
we first localize task-relevant functional elements in the input views and obtain their pixel-level segmentation masks.
We then back-project each posed RGB-D frame into 3D using known camera parameters to reconstruct the scene point cloud, following prior work~\cite{peng2023openscene, delitzas2024scenefun3d}.
The 2D segmentation masks of the functional elements are then back-projected and fused across views to produce 3D masks corresponding to the functional elements.

\paragraph{LLM-based contact graph reasoning.}
While the detected functional elements indicate \emph{what} scene components are relevant to the task, they do not specify \emph{how} the human body should interact with them, nor how the body is supported by the surrounding scene geometry (e.g., the floor).
To represent human-scene contact relations in a structured form, following prior work~\cite{hassan2019resolving, li2025genhsi}, we define a \textit{property graph}:
\begin{equation}
\mathcal{G} = (\mathcal{V}, \mathcal{E}), \quad 
\mathcal{V} = \mathcal{V}_{\text{body}} \cup \mathcal{V}_{\text{scene}},
\end{equation}
where $\mathcal{V}_{\text{body}}$ denotes a predefined set of SMPL-X body parts, and
$\mathcal{V}_{\text{scene}}$ denotes functional or supporting scene elements.
Each edge $(b, o) \in \mathcal{E}$ encodes a contact relation between a body part $b \in \mathcal{V}_{\text{body}}$ and a scene element $o \in \mathcal{V}_{\text{scene}}$.
Body-part names are annotated on the SMPL-X template (see Sup. Mat. Fig.~\ref{supmat:fig:contact_annot}) and are fixed across all experiments.
We then prompt a large language model (LLM), e.g., GPT-4o~\cite{OpenAI_ChatGPT} or Gemini, with the task description, the detected functional elements, the predefined body-part set, and additional structured instructions that encourage task-complete and human-like interactions.
The LLM outputs a contact graph $\mathcal{G}$, which specifies the involved body parts, the functional and supporting scene elements, and their corresponding contact relations (see Fig.~\ref{fig:framework}).
Similar to functional elements, inferred supporting elements (e.g., the floor) are segmented in each image and lifted to 3D masks.

\subsection{Functionality-aware Body Initialization}
\label{sec:4.2}
Although the inferred contact graph $\mathcal{G}$ provides high-level interaction constraints, directly fitting a 3D human body to the scene remains challenging due to the strong sensitivity of optimization-based methods to initialization.
To obtain a reliable initial body configuration, we first synthesize a human performing the task in the image and then estimate the corresponding 3D body and hand poses.
Since image-based synthesis may introduce left-right inconsistencies with the inferred contact graph, we update the contact graph to align its laterality with the initialized human body.

\begin{figure}[!t]
	\vspace{-4mm}
	\centering
	\includegraphics[width=\columnwidth]{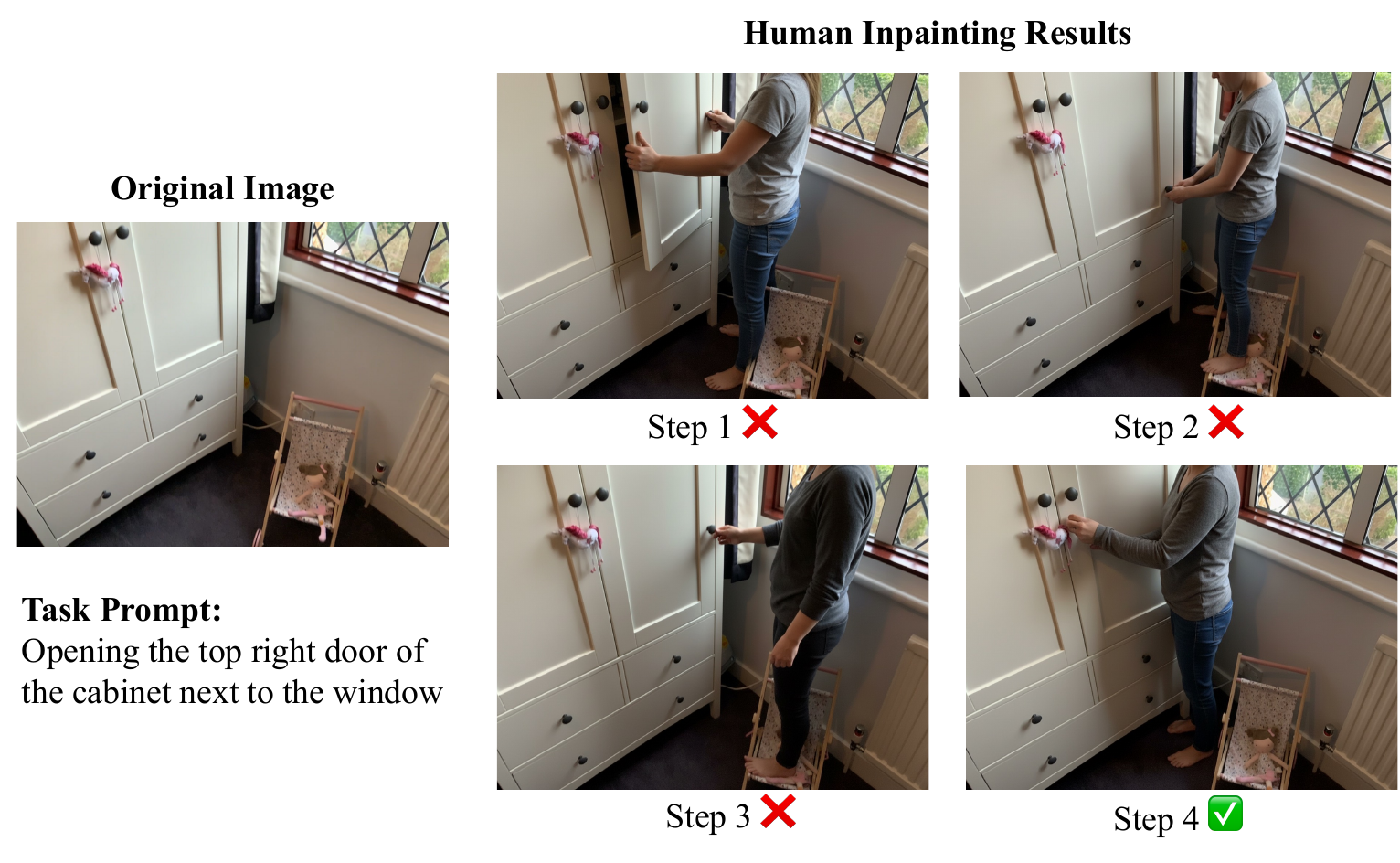}
	\vspace{-8mm}
	\caption{\textbf{Visualization of the human inpainting optimization process.} By automatically evaluating the human inpainting results, the image generation process is optimized to produce more reliable outcomes, thus strongly facilitating the subsequent body optimization step.}
	\label{fig:inpainting}
    \vspace{-2mm}
\end{figure}

\paragraph{Human inpainting with contact-aware reasoning.}
We employ a vision-language model (VLM), specifically Gemini~\citep{comanici2025gemini}, to synthesize human pixels in the input image.
To encourage the generated human to perform the specified task and establish appropriate contacts with the scene, we introduce a contact-aware prompting strategy.
In addition to the input image without humans and the task description, the inpainting prompt incorporates the inferred contact graph and the detected object bounding boxes.
These cues explicitly specify task-relevant functional and supporting elements, guiding the model to generate human body parts in spatial proximity to the target objects.
However, image inpainting models may hallucinate, unintentionally altering scene structures or introducing spurious objects, as illustrated in Fig.~\ref{fig:inpainting}.
To mitigate this issue, we adopt an iterative generator-critic scheme inspired by LLM-based optimization~\cite{yang2023large}.
A separate Gemini model is used as a critic to compare the inpainted image with the original input and verify that (1) the generated human performs the specified task, (2) contacts with functional elements are plausible, and (3) no irrelevant or non-existent objects are introduced.
If any criterion is violated, the generator is prompted to regenerate the human appearance.
This process is repeated until all criteria are satisfied or a maximum number of iterations is reached.
In practice, we find that 3-4 iterations are sufficient and outperform single-pass image generation.

\paragraph{3D human estimation.}
Given the human-inpainted image, we estimate SMPL-X parameters to initialize the 3D human body.
Specifically, we estimate the global translation $\mathbf{r}$, root orientation $\boldsymbol{\varphi}$, and body pose $\boldsymbol{\theta}^{b}$ using CameraHMR~\cite{patel2025camerahmr}, and estimate hand pose parameters $\boldsymbol{\theta}^{h}$ using WiLoR~\cite{potamias2025wilor}.
For cases where the hands are occluded in the image, $\boldsymbol{\theta}^{h}$ is initialized to the default relaxed hand pose of SMPL-X.
The estimated SMPL-X body is then transformed from the camera coordinate system to the world coordinate system using the known camera pose, ensuring that the human body and the scene are represented in a common reference frame.
The resulting SMPL-X parameters provide a task-specific and geometrically plausible initialization, which substantially simplifies the subsequent body refinement stage.

\paragraph{Contact graph refinement.}
We observe that image generation models may fail to consistently capture left-right spatial relations.
For example, as shown in Fig~\ref{fig:hand_swap}, the synthesized image may depict the left hand contacting a handle, even when the inferred contact graph specifies contact with the right hand.
Such laterality inconsistencies between the initialized body configuration and the contact graph can lead to invalid human-scene interactions during subsequent refinement.
To address this issue, we refine the contact graph by aligning its laterality with the inpainted image.
Specifically, we project the left and right wrist joints of the estimated 3D body onto the 2D image plane and compute their distances to the center $\mathbf{c}_o$ of the functional element bounding box:
\begin{equation}
d_{\text{left}} = \|\Pi(\mathbf{w}_{\text{left}}) - \mathbf{c}_o\|_2, \quad
d_{\text{right}} = \|\Pi(\mathbf{w}_{\text{right}}) - \mathbf{c}_o\|_2,
\end{equation}
where $\Pi(\cdot)$ denotes the 3D-to-2D projection operator and
$\mathbf{w}_{\text{left}}, \mathbf{w}_{\text{right}}$ are the 3D wrist joints.
If $d_{\text{left}} > d_{\text{right}} + \delta$, where $\delta$ is a small tolerance to account for projection noise and pose estimation errors, we apply a symmetric left-right swap to all hand-related nodes in the contact graph $\mathcal{G}$ (e.g., palm and finger nodes).
Otherwise, the contact graph remains unchanged.
This simple distance-based criterion is effective at resolving left-right ambiguities across different scenes and camera viewpoints.
The refined contact graph is denoted as $\mathcal{G}^*$.

\subsection{Optimization-based Body Refinement}
\label{sec:4.3}
To refine the body pose, global configurations, and the contact, a conventional solution is to jointly optimize all SMPL-X parameters.
However, we find in our trials that such joint optimization often leads to unrealistic HSI results, such as unnatural facing orientation and penetration to the scene. 
Therefore, we propose a two-stage coarse-to-fine optimization method to gradually refine the initial body state. 
This will not only preserve nuances in the initial body pose, but also improve the body-scene contact, making the 3D human body performing the specified task.

\paragraph{Optimization objective.}
To penalize body-scene interpenetration, we define a collision loss based on the signed distance field (SDF) of the SMPL-X body.
Given a scene point cloud $\mathcal{P} = \{\mathbf{p}_j\}_{j=1}^{N}$, the collision loss is formulated as
\begin{equation}
\mathcal{L}_{\text{col}}
=
\sum_{j=1}^{N}
\max\bigl(0,\; -\Psi(\mathbf{p}_j; \beta, r, \varphi, \theta)\bigr),
\end{equation}
where $\Psi(\mathbf{p}_j; \beta, r, \varphi, \theta)$ denotes the SDF value of point $\mathbf{p}_j$ with respect to the current SMPL-X body configuration, computed using VolumetricSMPL~\cite{mihajlovic2025volumetricsmpl}.
This loss penalizes scene points that lie inside the body volume and evaluates to zero when no interpenetration occurs.

To further enforce task-consistent body-scene contact, we introduce a contact loss guided by the refined contact graph $\mathcal{G}^*$.
For each contact pair $(b,o)\in\mathcal{G}^*$, where $b$ denotes a body part and $o$ a corresponding scene element, we minimize the distance between the body vertices $\mathcal{V}_b$ and the scene points $\mathcal{S}_o$ using a single-sided Chamfer distance:
\begin{equation}
\mathcal{L}_{\text{con}}
=
\sum_{(b,o)\in\mathcal{G}^*}
\frac{1}{|\mathcal{V}_b|}
\sum_{\mathbf{v}\in\mathcal{V}_b}
\min_{\mathbf{s}\in\mathcal{S}_o}
\|\mathbf{v}-\mathbf{s}\|_2^2 .
\end{equation}
The single-sided formulation pulls the body toward the intended contact surfaces without over-constraining the scene geometry.
For foot contacts, the loss is computed only on vertices near the toes and heel, allowing fine-grained poses such as tiptoe standing.
To regularize the pose space during optimization, we incorporate a VPoser prior~\cite{SMPL-X:2019}.
Specifically, we define
\begin{equation}
\mathcal{L}_{\text{prior}} = \|\,\mathbf{z}\,\|_2^2,
\qquad
\mathbf{z} = \mathrm{VPoserEnc}(\theta^{b}),
\end{equation}
where $\mathrm{VPoserEnc}(\cdot)$ denotes the VPoser encoder and $\mathbf{z}$ is encouraged to follow a standard normal distribution.
The overall optimization objective is defined as
\begin{equation}
\mathcal{L}
=
\lambda_{\text{col}}\,\mathcal{L}_{\text{col}}
+
\lambda_{\text{con}}\,\mathcal{L}_{\text{con}}
+
\lambda_{\text{prior}}\,\mathcal{L}_{\text{prior}},
\end{equation}
where $\lambda_{\text{col}}$, $\lambda_{\text{con}}$, and $\lambda_{\text{prior}}$ are scalar weighting coefficients.

\paragraph{Two-stage optimization strategy.}
As summarized in Algorithm~\ref{alg:two_stage_refine}, the refinement is carried out in two stages.
In the first stage, we optimize the 3D translation $r$, the global body orientation around the gravity axis $\varphi_g$, and the arm pose parameters $\theta^{\text{arm}}$.
Jointly optimizing the arm articulation and global translation enables the hands to reach and establish contact with the target functional elements specified by the task.
To preserve physical realism, the global orientation is restricted to rotations around the gravity axis, which prevents unnatural body tilting while still allowing feasible interaction configurations and obstacle avoidance.
The second stage focuses on improving physical plausibility and contact stability.
In this stage, we optimize the full body pose $\theta$ together with the 3D translation $r$, with particular emphasis on the ankle joints to ensure stable foot-ground contact.
A smaller learning rate $\eta_2$ (set to $\frac{1}{5}\eta_1$) is adopted to allow subtle pose adjustments without disrupting the refined configuration.
The pose prior loss $\mathcal{L}_{\text{prior}}$ is applied only in this stage to maintain anatomically valid body poses.

%%%%%%%%%%%%%%%%%%%%%%%%optimization algorithm%%%%%%%%%%%%%%%%%%%%%%%%
\begin{algorithm}[t]
\footnotesize
\SetAlgoLined
\KwIn{
Reconstructed scene point cloud $\mathcal{P}$;
refined contact graph $\mathcal{G}^*$;
initial SMPL-X parameters $(\beta, r_0, \varphi_0, \theta_0)$ (Sec.~4.2);
learning rates $\eta_1,\eta_2$;
iterations $K_1,K_2$;
loss weights $\lambda_{\text{col}},\lambda_{\text{con}},\lambda_{\text{prior}}$.
}
\KwOut{Refined SMPL-X parameters $(\beta, r^*, \varphi^*, \theta^*)$.}

\textbf{Initialization:}
$(r,\varphi,\theta)\leftarrow(r_0,\varphi_0,\theta_0)$.\;

\BlankLine
\textbf{Stage 1: Global alignment and functional interaction refinement}\;
\textbf{Optimize:} translation $r$, gravity-axis rotation $\varphi_g$, and arm pose $\theta^{\text{arm}}$.\;
\textbf{Freeze:} remaining pose parameters in $\theta$ and non-gravity components of $\varphi$.\;

\For{$k \leftarrow 1$ \KwTo $K_1$}{
    compute $\mathcal{L}_{\text{col}}$ using the body SDF and $\mathcal{P}$\;
    compute $\mathcal{L}_{\text{con}}$ guided by $\mathcal{G}^*$\;
    $\mathcal{L} \leftarrow \lambda_{\text{col}}\mathcal{L}_{\text{col}} + \lambda_{\text{con}}\mathcal{L}_{\text{con}}$\;
    $(r,\varphi_g,\theta^{\text{arm}}) \leftarrow (r,\varphi_g,\theta^{\text{arm}}) - \eta_1 \nabla \mathcal{L}$\;
}

\BlankLine
\textbf{Stage 2: Local pose refinement for physical stability}\;
\textbf{Optimize:} translation $r$ and full body pose $\theta$ (with emphasis on ankle joints).\;
\textbf{Freeze:} shape $\beta$ and non-gravity components of $\varphi$.\;

\For{$k \leftarrow 1$ \KwTo $K_2$}{
    compute $\mathcal{L}_{\text{col}}$ using the body SDF and $\mathcal{P}$\;
    compute $\mathcal{L}_{\text{con}}$ guided by $\mathcal{G}^*$\;
    compute $\mathcal{L}_{\text{prior}}$ using the VPoser prior\;
    $\mathcal{L} \leftarrow
    \lambda_{\text{col}}\mathcal{L}_{\text{col}}
    + \lambda_{\text{con}}\mathcal{L}_{\text{con}}
    + \lambda_{\text{prior}}\mathcal{L}_{\text{prior}}$\;
    $(r,\theta) \leftarrow (r,\theta) - \eta_2 \nabla \mathcal{L}$;
}

\Return $(\beta,r,\varphi,\theta)$\;
\caption{Two-stage optimization for refining SMPL-X body pose with collision avoidance and contact consistency.}
\label{alg:two_stage_refine}
\end{algorithm}

%%%%%%%%%%%%%%%%%%%%%%%%optimization algorithm%%%%%%%%%%%%%%%%%%%%%%%%

\section{Experiments}
\label{sec:exp}

\paragraph{Datasets.}
To evaluate both existing methods and our approach for human-scene interaction (HSI) synthesis, we construct a benchmark derived from the SceneFun3D dataset~\citep{delitzas2024scenefun3d}.
We select 30 indoor scenes with diverse layouts (living rooms, bedrooms, kitchens, and bathrooms), each containing three views with RGB images, depth maps, and mask annotations for key affordance elements (e.g., door handles, couches, and floors).
For each scene, we consider two evaluation settings: \emph{functional HSI} and \emph{general HSI}.
The \emph{functional HSI} prompts are taken from SceneFun3D and specify only the intended goal (e.g., \textit{open the door}, \textit{adjust the temperature}), requiring models to infer the relevant functional elements.
In contrast, \emph{general HSI} uses manually annotated prompts that explicitly describe both the action and the target object (e.g., \textit{sit on the chair}, \textit{stand in front of the window}).
This results in a total of 60 curated interaction tasks.
In addition, we capture real-world city scenes from multi-view images using GeoCalib~\cite{veicht2024geocalib} and MapAnything~\cite{keetha2025mapanything} to demonstrate compatibility with state-of-the-art feedforward 3D reconstruction pipelines.
Further details are provided in the supplementary material.

\begin{table}[t]
\centering
\resizebox{\columnwidth}{!}{%
\begin{tabular}{lcccc}
\toprule
\textbf{Method} &
\emph{SCS $\uparrow$} &
\emph{NCS $\uparrow$} &
\emph{N-FCD $\downarrow$} &
\emph{FCD $\downarrow$} \\
\midrule
\multicolumn{5}{c}{\textbf{General Human-scene Interaction}} \\ 
\midrule
GenZI*~\cite{li2024genzi} & \textbf{0.2542} & 0.9848 & 0.8496 & - \\
GenHSI*~\cite{li2025genhsi} & 0.2528 & 0.9906 & 0.7599 & - \\
\rowcolor[gray]{.9}\textbf{FunHSI (Ours)} & 0.2498 & \textbf{0.9929} & \textbf{0.7481} & - \\
\midrule
\multicolumn{5}{c}{\textbf{Functional Human-scene Interaction}} \\ 
\midrule
GenZI*~\cite{li2024genzi} & 0.2501 & 0.9823 & 0.2027 & 0.6262 \\
GenHSI*~\cite{li2025genhsi} & \textbf{0.2607} & \textbf{0.9925} & 0.5415 & 0.4199 \\
\rowcolor[gray]{.9}\textbf{FunHSI (Ours)} & 0.2540 & 0.9917 & \textbf{0.1837} & \textbf{0.2968} \\
\bottomrule
\end{tabular}}
\caption{
\textbf{Quantitative Comparison on our curated SceneFun3d subset.}
Best scores are in boldface. The symbol * denotes that the baselines are their modified versions for fair comparison.
}
\label{tab:comparison}
\end{table}

\paragraph{Evaluation Metrics.}
We evaluate HSI synthesis using 4 complementary metrics, i.e. \emph{semantic consistency score (SCS)}, \emph{non-collision score (NCS)}, \emph{non-functional contact distance (N-FCD)}, and \emph{functional contact distance (FCD)}, respectively.
The \emph{semantic consistency score} measures the alignment between the synthesized 3D interaction and the input text prompt.
We compute a CLIP score~\cite{radford2021learning} by rendering each synthesized interaction into three views, extracting image-text cosine similarities using CLIP ViT-B/32, and averaging the scores across views.
For \emph{non-collision score}, we compute a non-collision score based on penetration between the SMPL-X body mesh and the reconstructed scene point cloud, following VolumetricSMPL~\cite{mihajlovic2025volumetricsmpl}.
For \emph{non-functional contact distance}, we use the average Chamfer distance between the human body mesh and supporting scene elements (e.g., the floor or chair).
The \emph{functional contact distance} assesses whether the synthesized interaction has appropriate contact with task-relevant functional elements, e.g., a hand touching a door handle in the task of ``open the door''.
This metric is computed as the Chamfer distance between the functional element region and the interacting human hands. 

\paragraph{Baselines.}
To our knowledge, no existing method explicitly targets \emph{functional} human-scene interactions in 3D.
We therefore compare our approach with the most closely related baselines.
GenZI~\citep{li2024genzi} synthesizes human appearances in individual views and reconstructs a 3D body via multi-view fitting.
For a fair comparison, we adapt GenZI to operate on the same three posed RGB-D images used in our benchmark.
GenHSI~\citep{li2025genhsi} proposes a training-free pipeline for generating long human-scene interaction videos by combining keyframe planning, 3D-aware inpainting, and motion animation.
We extend GenHSI with functional element detection, perform human inpainting from randomly sampled views, and apply its original body-fitting strategy to our inputs.
Due to these adaptations, the resulting baselines are denoted as GenZI* and GenHSI*, respectively.

\begin{figure}[!t]
	\centering
    \vspace{-4mm}
	\includegraphics[width=\columnwidth]{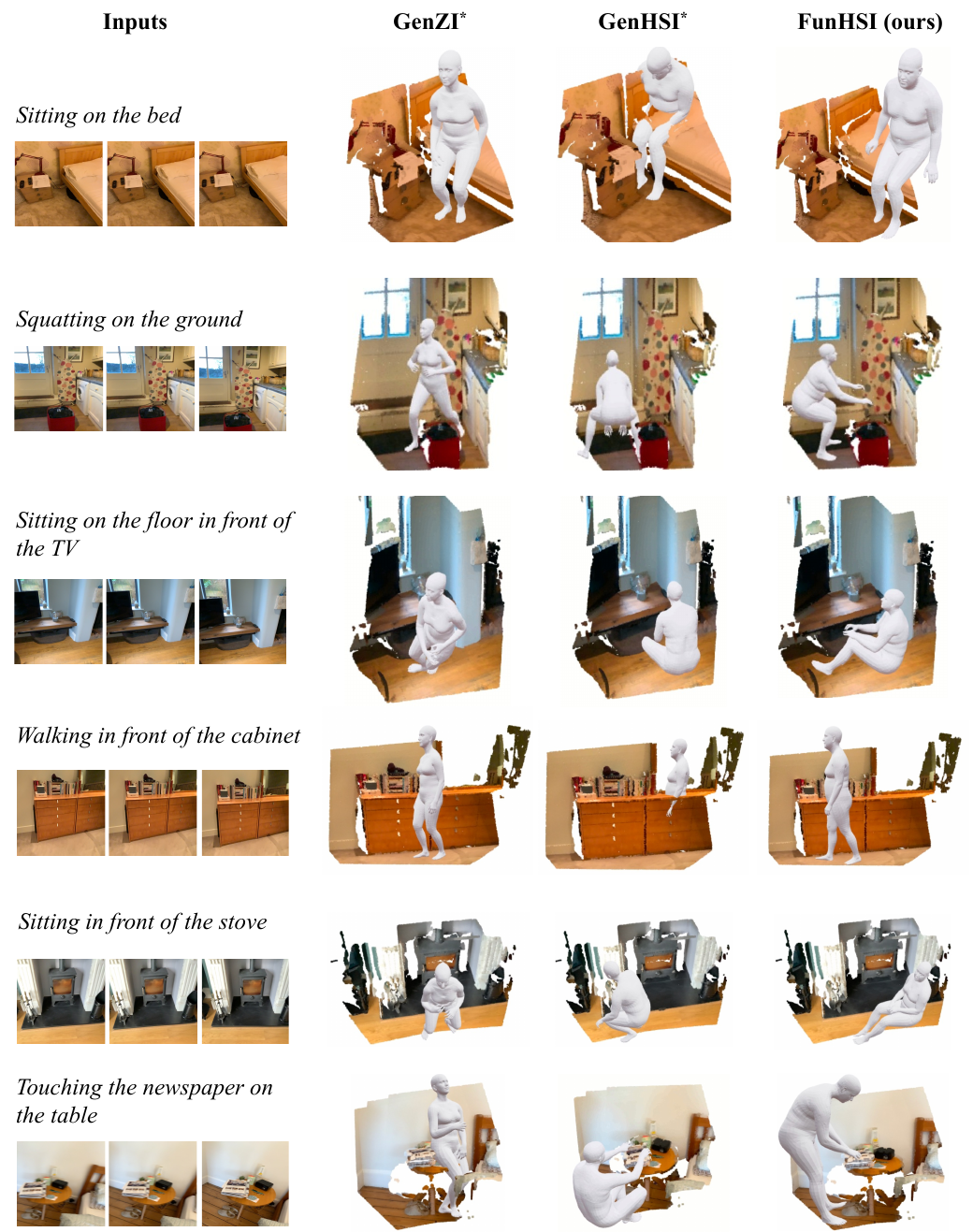}
	\vspace{-6mm}
    \caption{\textbf{Qualitative results on SceneFun3D for general human-scene interaction.}
    We compare GenZI*, GenHSI*, and our FunHSI with non-functional prompts such as sitting, squatting, and walking.}
    \vspace{-2mm}
	\label{fig:qualitative_hsi}
\end{figure} 

\begin{figure*}[!t]
	\centering
	\includegraphics[width=0.95\linewidth]{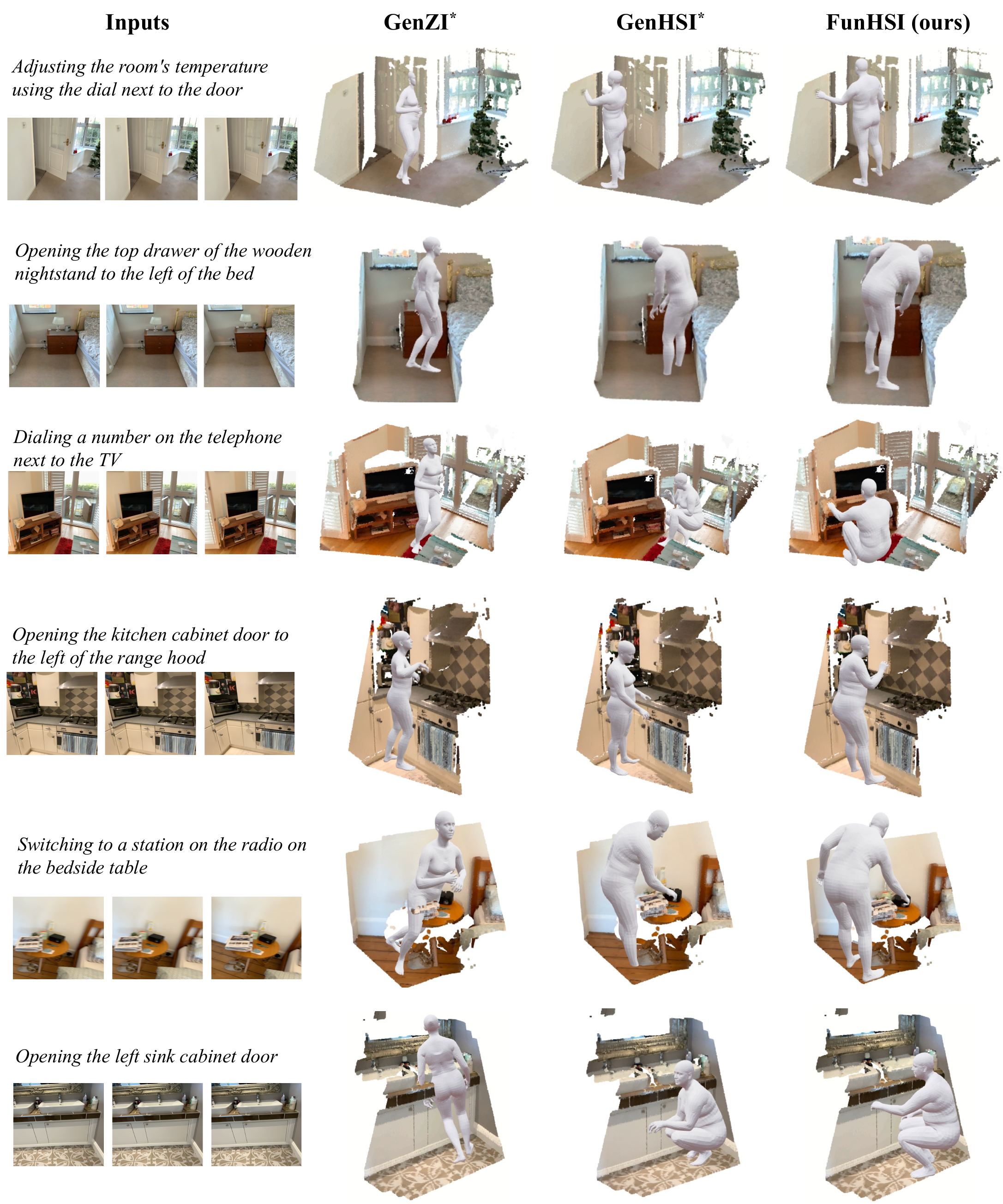}
	\vspace{-4mm}
    \caption{\textbf{Qualitative results on SceneFun3D for functional human-scene interaction.}
    Given open-vocabulary functional commands (e.g., adjusting temperature, dialing a number, switching a radio station) and posed RGB-D inputs, we compare GenZI*, GenHSI*, and our FunHSI.
    Existing methods struggle to reason about task intent and often interact with incorrect objects or miss fine-grained functional components.
    In contrast, FunHSI accurately identifies task-relevant functional elements and generates physically plausible 3D human poses that establish correct contacts with both large objects and small functional parts (e.g., knobs, dials, cabinet handles), demonstrating robust functional grounding and contact reasoning.}
	\label{fig:qualitative}
\end{figure*} 

\subsection{Comparison to Baselines}
\paragraph{Quantitative Evaluation.}
Table~\ref{tab:comparison} summarizes the quantitative comparison between our FunHSI method and the modified baselines.
Overall, FunHSI performs competitively in the \emph{general HSI} setting and substantially outperforms the baselines in \emph{functional HSI}.
For general HSI, FunHSI achieves comparable semantic consistency (0.2498) while improving physical plausibility.
In particular, it attains the lowest contact distance (0.7481), outperforming GenZI* (0.8496) and GenHSI* (0.7599), together with a slightly higher non-collision score (0.9929), indicating that improved contact quality is not achieved at the cost of increased body-scene penetration.
For functional HSI, FunHSI consistently yields the best results, with the lowest functional contact distance (0.2968) and the lowest overall contact distance (0.1837), significantly outperforming GenZI* and GenHSI*.
Although GenHSI* achieves a marginally higher non-collision score (0.9925 vs.\ 0.9917), FunHSI maintains comparable physical plausibility and semantic consistency (0.2540).

\begin{figure}[!t]
	\centering
	\includegraphics[width=\columnwidth]{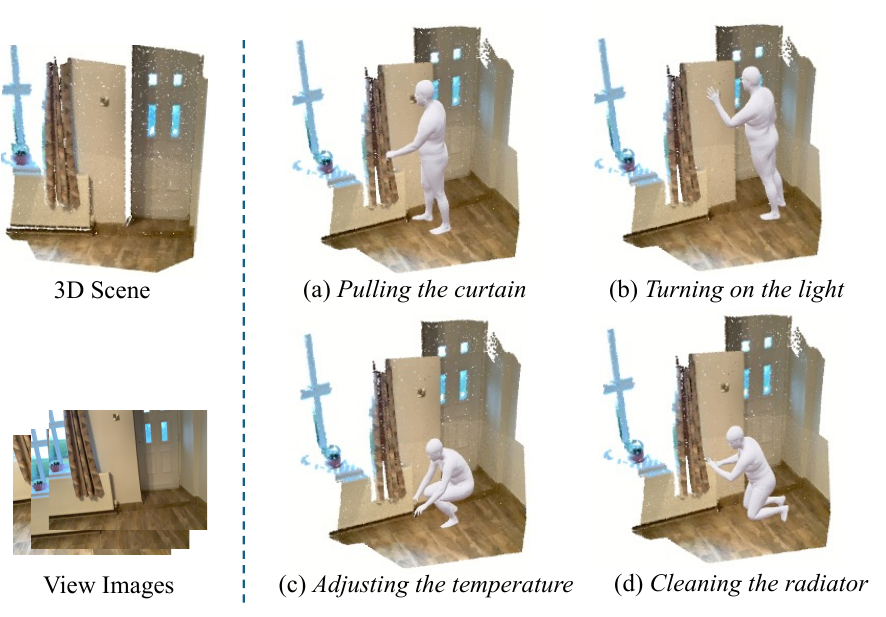}
	\vspace{-8mm}
    \caption{\textbf{Illustration of functionality awareness of FunHSI.}
    Given the same 3D scene, FunHSI generates distinct human-scene interactions conditioned on different high-level task prompts.}
    \vspace{-2mm}
	\label{fig:fun_awareness}
\end{figure}

\begin{figure}[!t]
	\centering
	\includegraphics[width=0.96\columnwidth]{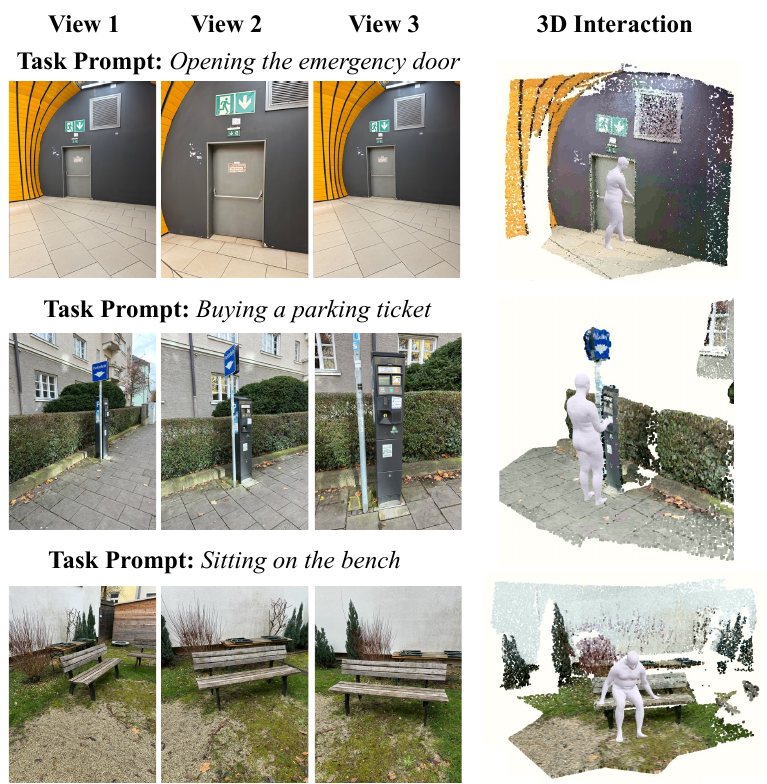}
	\vspace{-2mm}
    \caption{\textbf{Qualitative results on in-the-wild scenes.}
    We show our FunHSI results on real-world scenes captured by smart phone in Munich.}
    \vspace{-4mm}
	\label{fig:in_the_wild}
\end{figure}

\paragraph{Qualitative Evaluation.}
Fig.~\ref{fig:qualitative_hsi} and Fig.~\ref{fig:qualitative} show qualitative comparisons under both general and functional human-scene interaction scenarios.
For functional tasks that require identifying and interacting with task-relevant elements (e.g., operating knobs, opening drawers, or interacting with small appliances), the baseline methods often fail to localize the correct functional targets or produce inaccurate hand-object contacts.
In contrast, FunHSI consistently grounds interactions on the appropriate functional elements and generates realistic human-scene interactions.
For general interaction prompts such as sitting, squatting, or standing near scene objects, FunHSI produces perceptually plausible body poses and interactions, achieving performance comparable to the baseline methods.
Fig.~\ref{fig:fun_awareness} further illustrates the functional awareness of FunHSI: given different high-level task prompts abouth the same scene or object, the generated bodies accomplish the intended tasks with diverse and appropriate poses.
Additional visual results are provided in the supplementary material.

\paragraph{Generalization to City Scenes.}
Fig.~\ref{fig:in_the_wild} presents qualitative results on \emph{in-the-wild} city scenes captured using a smartphone in public spaces in a city.
Given multi-view RGB images reconstructed into 3D scenes, FunHSI successfully generates plausible human-scene interactions for diverse real-world tasks, such as opening an emergency door, buying a parking ticket, and sitting on a bench.
Despite the challenges posed by cluttered environments, noisy geometry, and incomplete reconstructions, our method robustly grounds interactions to the correct functional elements and produces physically plausible body poses.
These results demonstrate that FunHSI generalizes beyond curated indoor datasets and is compatible with real-world feedforward 3D reconstruction pipelines.
More visualizations are provided in Fig.~\ref{supmat:fig:app-wild} of the supplementary material.

\begin{figure}[!t]
	\centering
	\includegraphics[width=0.9\columnwidth]{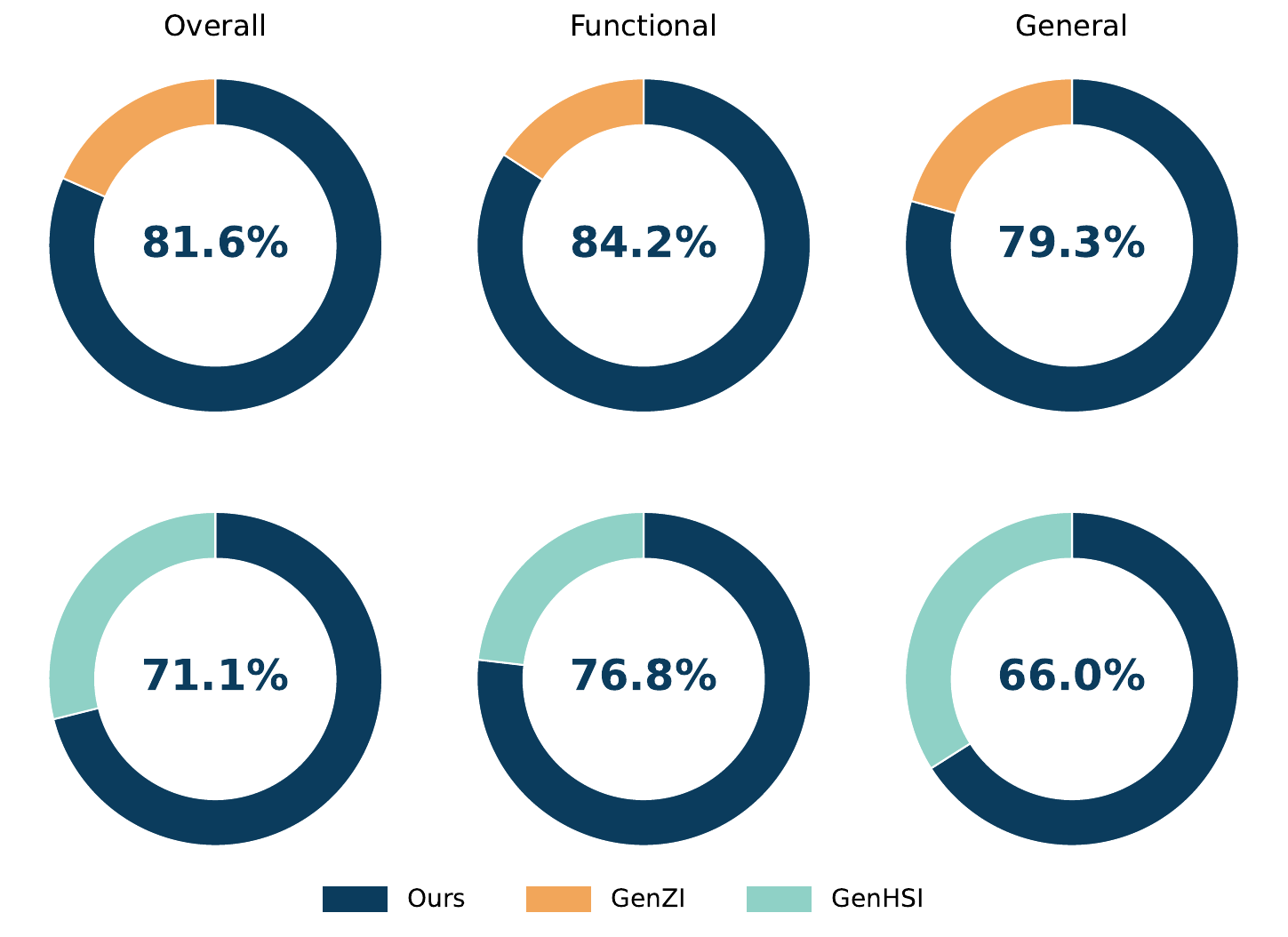}
	\vspace{-4mm}
    \caption{\textbf{User study of 3D human–scene interaction synthesis on our curated dataset.} Participants show a strong preference for our method over baselines (i.e., GenHSI~\cite{li2025genhsi} and GenZI~\cite{li2024genzi}) under both functional and general HSI settings.}
	\label{fig:user_study}
    \vspace{-2mm}
\end{figure} 

\subsection{Perceptual User Study}
\label{sec:user_study}
We conduct a perceptual user study to evaluate the visual quality and interaction realism of synthesized 3D human--scene interactions.
The study is performed on the SceneFun3D benchmark under both \emph{functional HSI} and \emph{general HSI} settings.
Participants are presented with rendered interaction results generated by FunHSI and the baseline methods, and are asked to select the most plausible and realistic human--scene interaction for each task.
The evaluation focuses on overall perceptual quality, including the appropriateness of body pose, physical plausibility of contact, and consistency with the given task prompt.
Fig.~\ref{fig:user_study} summarizes the user preference results.
Overall, FunHSI is strongly preferred over the baseline methods across all evaluation settings.
When taking GenHSI as a representative baseline, FunHSI achieves an overall preference rate of 71.1\%.
When evaluated separately, FunHSI obtains a preference rate of 76.8\% for functional HSI and 66.0\% for general HSI, indicating a clear advantage in scenarios that require functional reasoning and affordance-aware interaction.
Moreover, the preference margins are more pronounced in functional HSI, indicating that users are particularly sensitive to correct functional grounding and realistic contact with task-relevant elements.
These results demonstrate that FunHSI not only improves quantitative metrics, but also produces perceptually more convincing human--scene interactions.

\begin{table}[t]
\centering
\resizebox{\columnwidth}{!}{%
\begin{tabular}{lccc}
\toprule
\textbf{Method} &  
\emph{NCS $\uparrow$} &
\emph{N-FCD $\downarrow$} & 
\emph{FCD $\downarrow$} \\
\midrule
w/o contact graph refinement & 0.9913 & 0.2892 & 0.2962 \\
w/o body \& hand estimation & 0.9889 & 0.2956 & 0.4724 \\
w/o iterative body refinement & 0.9798 & 0.6067 & 0.6561 \\
\rowcolor[gray]{.9}\textbf{FunHSI} & 0.9917 & \textbf{0.1837} & 0.2968 \\
\rowcolor[gray]{.9}\textbf{FunHSI} + oracle detection & \textbf{0.9918} & 0.2155 & \textbf{0.2662} \\
\bottomrule
\end{tabular}}
\caption{\textbf{Ablation study of key components on our curated dataset.}
Each component contributes to the overall performance, and using oracle detection further improves the results.
}
\vspace{-2mm}
\label{tab:ablation}
\end{table}

\begin{figure}[!t]
	\centering
	\includegraphics[width=\columnwidth]{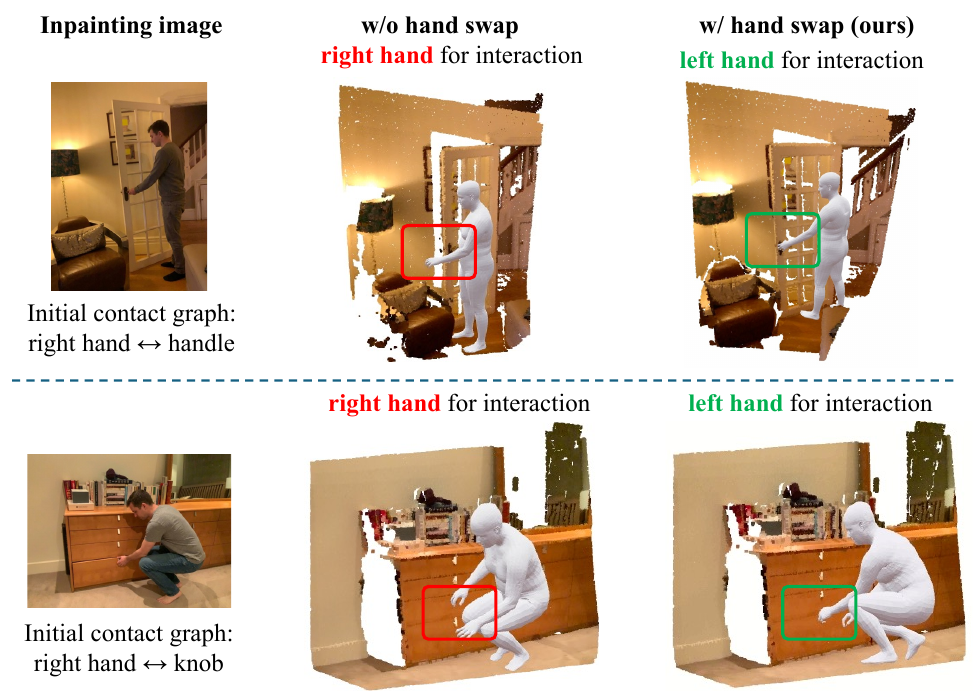}
	\vspace{-4mm}
    \caption{\textbf{Illustration of resolving left-right hand ambiguity via contact graph refinement.}
    Directly enforcing initial contact graphs results in unnatural or physically implausible interactions (red).
    By swapping left-right hand to align with the observed contacting hand in the image, our method produces correct and stable human-scene interactions (green).}
    \vspace{-2mm}
	\label{fig:hand_swap}
\end{figure}

\subsection{Ablation Studies}
\paragraph{Contact graph refinement.}
We ablate the contact graph refinement module by directly using the initial contact graph predicted by the LLM, without aligning left-right relations to the inpainting image.
As shown in Table~\ref{tab:ablation} and Fig.~\ref{fig:hand_swap}, removing this refinement leads to degraded contact accuracy, particularly for supporting elements such as the floor, while only marginally affecting the functional distance.
This behavior indicates that ambiguities in left-right correspondence between the contact graph and the generated image can cause failures in the body fitting stage, highlighting the importance of contact graph refinement for stable and accurate interactions.

\begin{figure}[!t]
	\centering
	\includegraphics[width=\columnwidth]{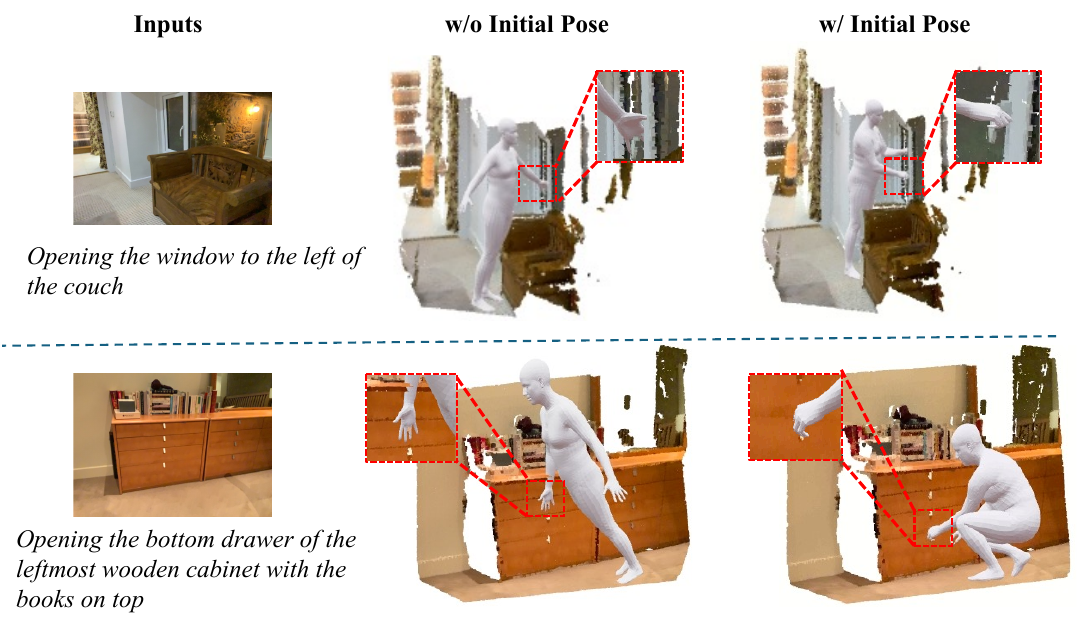}
	\vspace{-8mm}
    \caption{\textbf{Effect of body and hand pose initialization.}
     Body and hand pose initialization provides a consistent starting point, enabling correct hand placement and stable refinement for functional interactions.}
     \vspace{-2mm}
	\label{fig:init_pose}
\end{figure}

\paragraph{Body \& hand pose estimation.}
We evaluate the importance of body and hand pose estimation by removing this module from our pipeline and initializing the SMPL-X body with a T-pose prior to refinement.
As shown in Table~\ref{tab:ablation} and Fig.~\ref{fig:init_pose}, this modification leads to consistent degradation across all metrics.
This observation indicates that accurate body and hand pose estimation from the inpainted image plays a critical role in guiding the optimization.

\paragraph{Body refinement.}
We ablate the body refinement stage by directly using the estimated SMPL-X pose without further optimization.
As shown in Table~\ref{tab:ablation}, this results in increased body-scene penetration and less realistic contacts, indicating that the initial pose alone is insufficient to resolve geometric inconsistencies.
These results confirm the necessity of body refinement for producing physically plausible and functionally correct HSI.

\paragraph{Functional element detection.} 
To evaluate the impact of detection accuracy, we replace the predicted functional element masks with ground-truth annotations (i.e., oracle detection). 
As reported in Table~\ref{tab:ablation}, oracle detection leads to a noticeable reduction in functional contact distance (from 0.2968 to 0.2662) while preserving comparable non-collision performance. 
This improvement suggests that our generation framework can directly benefit from more robust upstream detection modules.

\section{Conclusion}
\label{sec:con}
In this work, we studied the problem of functional human-scene interaction synthesis, where a human must reason about object functionality and establish appropriate physical contact to accomplish an open-vocabulary task in a novel 3D scene.
We proposed FunHSI, a training-free and functionality-driven framework that generates 3D human-scene interactions from posed RGB-D observations and open-vocabulary task prompts, without relying on explicit action-object descriptions.
By integrating functionality-aware contact graph reasoning, human initialization, and optimization-based body refinement, FunHSI bridges high-level task intent and physically plausible interaction.
Extensive evaluations on a benchmark derived from SceneFun3D show that FunHSI consistently outperforms existing baselines, particularly for functional interactions, while maintaining strong physical plausibility.
We believe FunHSI represents a step toward more semantically grounded human-scene interaction synthesis and opens up future directions for long-horizon and real-world embodied interaction.

\paragraph{Limitations and future work.}
Our method currently focuses on single-step functional human-scene interactions, where a single human pose is synthesized to accomplish a given task.
As a result, it does not explicitly model long-horizon or multi-step interactions that require sequential planning or temporal reasoning across multiple actions (e.g., opening a door and then walking through it).
Extending FunHSI to support temporally coherent, multi-step functional interactions remains an interesting direction for future work.
In addition, the scales of city scenes are estimated from RGB images. Unifying the body and the scene scales is also a future work.

\section*{Acknowledgement}
We sincerely thank Alexandros Delitzas and Francis Engelmann for the guidance on SceneFun3D, Priyanka Patel on the guidance of CameraHMR, Muhammed Kocabas for fruitful discussions on foundation models. 
We also sincerely thank Nitin Saini and Nathan Bajandas for kind help and explorations on Unreal Engine. This work was done when Jie Liu was an intern at Meshcapade.
\paragraph{Disclosure.} While MJB is a co-founder and Chief Scientist at Meshcapade, his research in this project was performed solely at, and funded solely by, the Max Planck Society.

{
    \small
    \bibliographystyle{ieeenat_fullname}
    \bibliography{main}
}

\clearpage
\setcounter{page}{1}
\maketitlesupplementary
\appendix

\begin{figure}[!t]
	\vspace{-4mm}
	\centering
	\includegraphics[width=\linewidth]{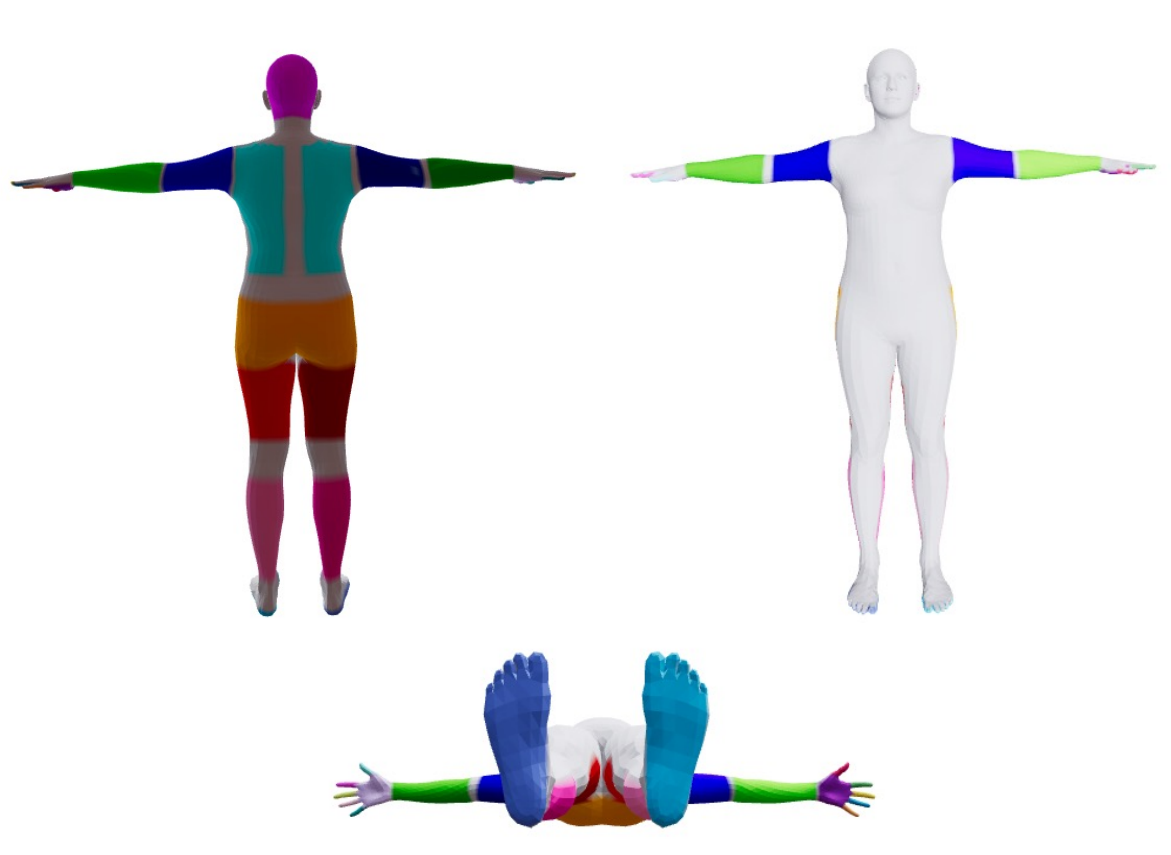}
	\vspace{-4mm}
	\caption{\textbf{Illustrations of human body part annotation.}
    The SMPL-X body surface is decomposed into semantically meaningful parts for contact reasoning and body refinement.
    Hands are further subdivided into the palm and individual fingers, enabling fine-grained modeling of hand–object interactions with small functional elements.}
	\label{supmat:fig:contact_annot}
\end{figure} 

\begin{figure*}[!t]
	\centering
	\includegraphics[width=\linewidth]{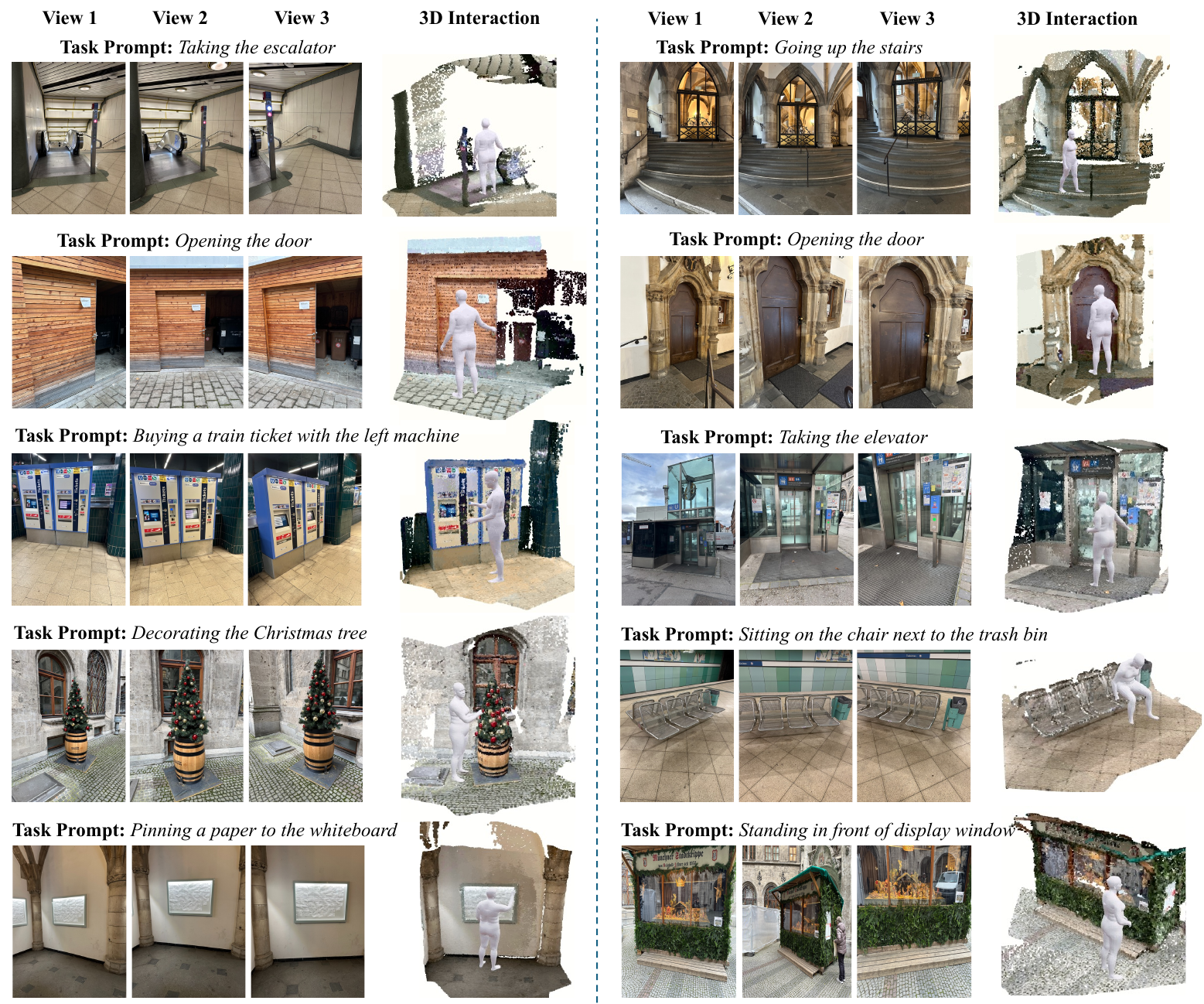}
    \caption{\textbf{Additional qualitative results on real-world scenes.}
    Given three posed RGB-D views and a high-level task prompt, FunHSI generates both functionality-aware and general human-scene interactions in diverse public environments.
    The examples cover a wide range of actions, including operating escalators and elevators, purchasing tickets, opening doors, and interacting with urban objects, highlighting the robustness and generalization ability of our method beyond indoor scenes.}
	\label{supmat:fig:app-wild}
\end{figure*} 

\begin{figure*}[!t]
	\centering
	\includegraphics[width=0.95\linewidth]{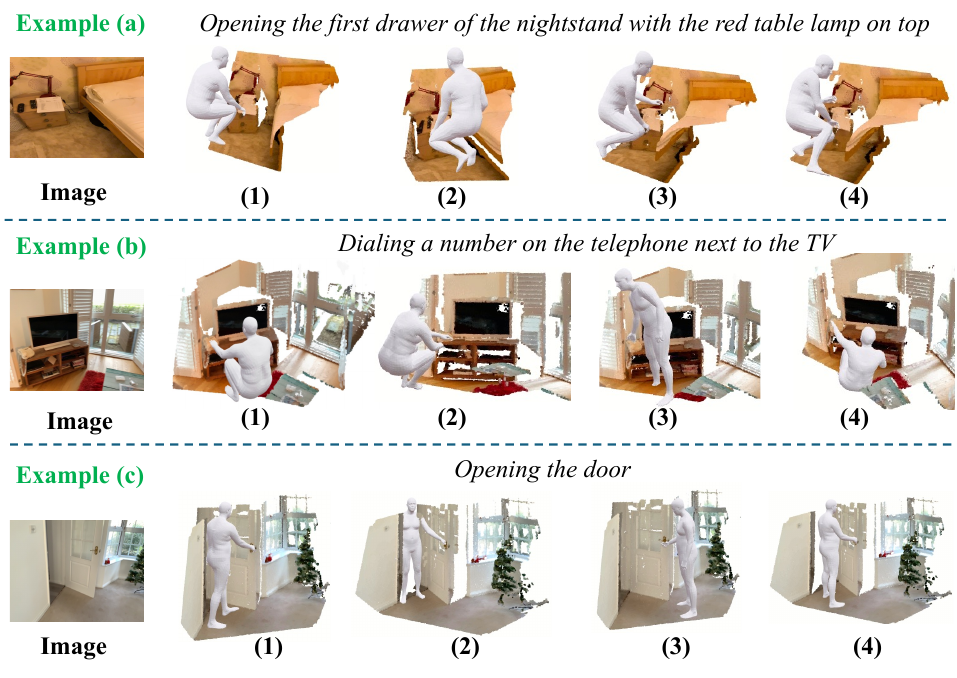}
    \caption{\textbf{Diversity of functionally consistent human-scene interactions generated by FunHSI.}
    For each example, we show multiple valid 3D human poses synthesized under the same scene and task prompt.
    Although the interactions differ in body configuration and spatial arrangement, all results preserve the intended functional contact, such as opening a drawer, dialing a telephone, or opening a door.
    This demonstrates that FunHSI supports diverse yet functionally consistent human-scene interaction generation.}
	\label{supmat:fig:diversity}
\end{figure*} 

\begin{figure*}[!t]
	\centering
	\includegraphics[width=0.95\linewidth]{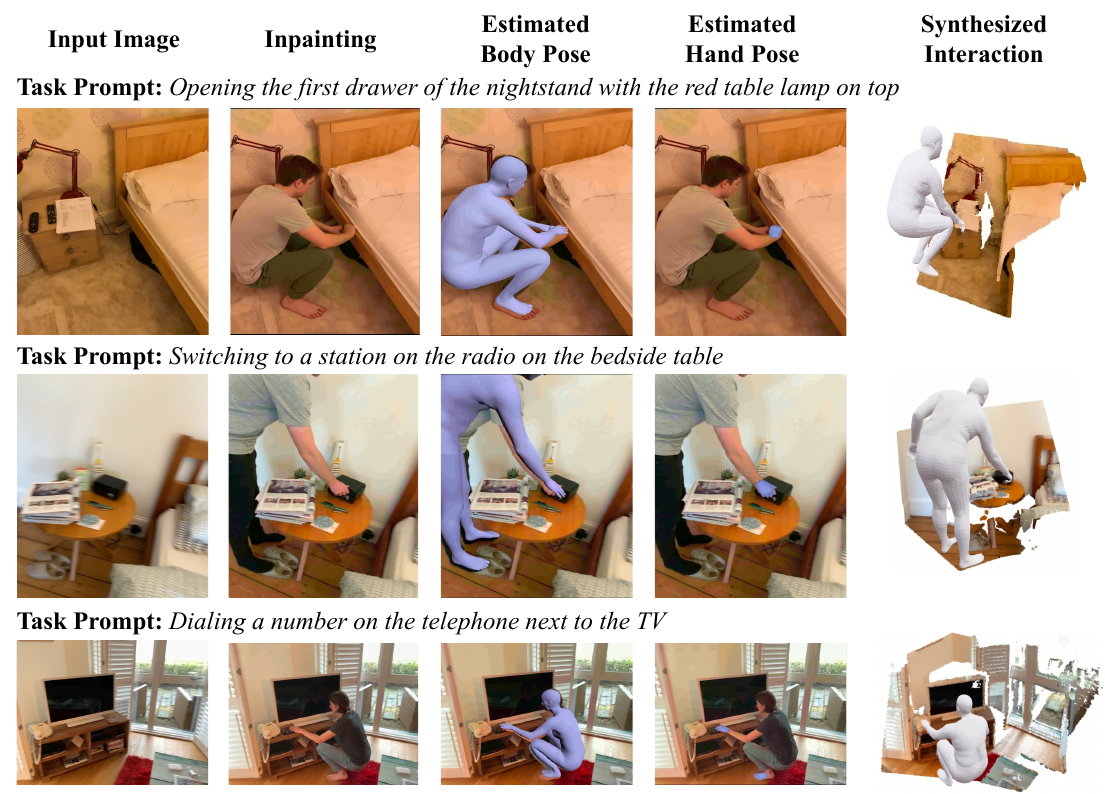}
    \caption{\textbf{Examples of human inpainting and pose estimation.}
    Given an input image, we first synthesize a task-consistent human via image inpainting.
    The inpainted result is then used to estimate the initial 3D body pose and articulated hand pose, which together serve as initialization for our subsequent 3D body refinement and interaction synthesis.}
	\label{supmat:fig:inpainting}
\end{figure*}

\section{Human Body Part Annotation}
To enable faithful, interpretable, and executable contact reasoning, we annotate the SMPL-X body surface using a hierarchical part decomposition. 
At the coarse level, we partition the body surface into 15 semantic parts following the SMPL-X template~\cite{SMPL-X:2019}:
\emph{head}, \emph{left upper arm}, \emph{right upper arm}, \emph{left forearm}, \emph{right forearm},
\emph{left hand}, \emph{right hand}, \emph{back}, \emph{buttocks},
\emph{left thigh}, \emph{right thigh}, \emph{left calf}, \emph{right calf}, \emph{left foot}, and \emph{right foot},
as illustrated in Fig.~\ref{supmat:fig:contact_annot}.
Each part corresponds to a fixed subset of vertices on the SMPL-X mesh, yielding consistent semantic labeling across different poses and body shapes.
Since functional interactions in indoor environments are primarily performed by the hands and often involve small-scale objects (e.g., knobs, switches, dials), we further introduce a fine-grained hand annotation.
Specifically, each hand is subdivided into six sub-parts: one \emph{palm} and five \emph{fingers}.
Each sub-part is associated with a predefined vertex set on the SMPL-X mesh, as shown in Fig.~\ref{supmat:fig:contact_annot}.
This design allows the representation of both whole-hand contacts (e.g., palm-handle) and finger-level functional interactions (e.g., index finger-button) without introducing unnecessary anatomical complexity.
This hierarchical annotation plays a dual role in our pipeline.
First, it provides a structured and semantically grounded vocabulary for LLM-based contact graph reasoning, enabling the model to express contacts using interpretable body-part names (e.g., ``left index finger touches the switch'').
Second, it establishes a direct mapping from contact semantics to geometric constraints: each contact node $b$ in the contact graph is mapped to its corresponding vertex set $\mathcal{V}_b$, which is used to compute contact losses during body refinement.
By grounding language-level contact reasoning in mesh-level geometry, this annotation enables precise functional interactions while maintaining physical plausibility.

\section{Datasets Details}
\paragraph{Indoor scenes from SceneFun3D~\cite{delitzas2024scenefun3d}.}
To systematically evaluate both prior methods and our approach for human-scene interaction (HSI) synthesis under fair and controlled settings, we construct a new benchmark derived from the SceneFun3D dataset.
We select 30 indoor scenes covering diverse spatial layouts and functional contexts, including living rooms, bedrooms, kitchens, and bathrooms.
All scenes contain common household objects that afford human interaction, such as doors, drawers, cabinets, switches, radiators, and supporting furniture.
For each scene, we provide three canonical RGB-D views captured from different viewpoints, where each view consists of an RGB image, a depth image, and pixel-level mask annotations for key affordance-bearing elements (e.g., door handles, knobs, floors, and supporting surfaces).
Using known camera parameters, the three views are back-projected and fused into a unified 3D point cloud, which serves as the geometric input for all interaction synthesis methods.
For each scene, we manually define two types of interaction settings: functional human-scene interaction (functional HSI) and non-functional human-scene interaction (general HSI).
Functional HSI requires the human to interact with a specific functional element to accomplish a task objective (e.g., \textit{open the door}, \textit{adjust the room temperature}, \textit{dial a number on the telephone}), while non-functional HSI involves generic body-scene interactions that do not rely on object functionality (e.g., \textit{sit on the floor}, \textit{stand in front of the window}).
Each interaction setting is paired with a single text prompt per scene, resulting in a total of 60 curated interaction tasks (30 functional and 30 non-functional).
The functional interaction prompts are designed to cover a diverse range of manipulation affordances, including \textit{pinch\_pull}, \textit{hook\_pull}, \textit{tip\_push}, \textit{rotate}, \textit{plug\_in}, \textit{unplug}, and \textit{key\_press}.
Most tasks involve fine-grained hand-object interactions, intentionally emphasizing functional reasoning and precise contact modeling rather than coarse body placement alone.
All methods are evaluated on the same set of scenes, views, and text prompts without additional training or scene-specific tuning.
The reconstructed scene geometry and affordance annotations are reused across different interaction prompts within each scene to ensure consistent and fair comparison.

\paragraph{Real-world city scenes.}
To evaluate the generalization ability of FunHSI under open-world conditions, we additionally collect a set of real-world city scenes captured in public environments.
All data are captured using an iPhone 14 Pro Max. For each scene, we take multiple RGB images from different viewpoints. We use GeoCalib~\cite{veicht2024geocalib} to estimate the camera intrinsic parameters and the gravity direction, and use MapAnything~\cite{keetha2025mapanything} to estimate the camera poses, the depth maps, and the 3D scene point cloud.

The collected scenes include diverse outdoor and semi-outdoor environments such as building entrances, staircases, ticket machines, escalators, benches, and public facilities, featuring challenging factors including clutter, reflective surfaces, varying illumination, and unconstrained object layouts.
We apply the same processing pipeline as in indoor scenes without any scene-specific tuning.
This experimental setting allows us to assess whether FunHSI can generalize beyond curated indoor datasets and reliably synthesize function-aware human-scene interactions in real-world, unconstrained environments.

\section{Implementation Details}
All our experiments are conducted on a single NVIDIA A6000 GPU.
For functionality grounding and contact reasoning, we use Gemini-2.5-Flash for functional element identification, Gemini Robotics-ER-1.5 for bounding box localization, and GPT-4o for contact graph generation.
All vision-language model queries are performed in a zero-shot manner, without task-specific fine-tuning.
Scene reconstruction is performed by back-projecting three posed RGB-D views into a unified point cloud using known camera parameters.
Functional and supporting elements are segmented using SAM-ViT-H and lifted into 3D.
The reconstructed scene geometry and functional element annotations are cached and reused across different interaction prompts within the same scene.
Human body initialization is obtained via image-space human inpainting using Gemini.
To reduce hallucinations, we apply a generator-evaluator loop with at most four iterations.
Initial 3D human parameters are estimated using CameraHMR~\cite{patel2025camerahmr} for body pose and WiLoR~\cite{potamias2025wilor} for hand pose.
For occluded hands, we initialize the hand pose using the relaxed SMPL-X default configuration.
Body refinement is performed using the two-stage optimization procedure described in Algorithm~\ref{alg:two_stage_refine}.
We use the AdamW optimizer for both stages.
In Stage~1, we optimize the 3D translation, gravity-axis global rotation, and arm pose parameters for $K_1=400$ iterations with learning rate $\eta_1 = 1\times10^{-2}$.
In Stage~2, we optimize the full body pose and translation for $K_2=200$ iterations using a reduced learning rate $\eta_2 = \eta_1/5$, together with the VPoser prior.
Unless otherwise specified, all hyperparameters are fixed across scenes and prompts.

\section{More Experimental Analysis}
\label{sec:results}
\paragraph{Additional results on real-world cenes.}
Fig.~\ref{supmat:fig:app-wild} presents additional qualitative results of our FunHSI on real-world scenes captured in public environments.
These scenes exhibit significantly higher visual and geometric complexity than indoor datasets, including cluttered backgrounds, irregular lighting conditions, reflective surfaces, and diverse object appearances.
Given three posed RGB-D views and a task-level text prompt, FunHSI successfully synthesizes functionally appropriate human-scene interactions without scene-specific tuning.
As shown in the figure, our method correctly identifies task-relevant functional elements and generates plausible interactions for a wide range of actions, such as taking escalators or elevators, buying tickets from vending machines, opening doors, pinning objects to a whiteboard, and interacting with urban furniture.
These results demonstrate that FunHSI generalizes beyond curated indoor datasets and is capable of handling open-world scenes while preserving functional grounding, contact correctness, and physical plausibility.

\paragraph{Generation Diversity.}
Fig.~\ref{supmat:fig:diversity} illustrates the diversity of human-scene interactions generated by FunHSI under the same scene and task prompt.
For each example, we visualize multiple valid 3D human poses that differ in body configuration, viewpoint, and spatial arrangement, while consistently preserving the intended functional contact.
Specifically, FunHSI produces diverse interaction realizations for tasks such as opening a drawer, dialing a telephone, and opening a door, all of which maintain correct contact with the task-relevant functional elements.
These variations arise from differences in initial image synthesis and subsequent geometric refinement, rather than changes in task specification.
This result demonstrates that FunHSI does not collapse to a single canonical pose, but instead supports diverse yet functionally consistent human-scene interaction generation.

\paragraph{Human Inpainting Examples.}
Fig.~\ref{fig:inpainting} presents representative examples of task-conditioned human inpainting in our pipeline.
Given an input RGB image and a task-level functional prompt, the inpainting model synthesizes a human that is spatially consistent with the scene layout and roughly aligned with the intended interaction region.
Importantly, the inpainted humans already reflect coarse functional intent (e.g., reaching, crouching, or bending), providing a semantically meaningful and visually grounded initialization that reduces ambiguity in subsequent 3D reconstruction.

\paragraph{Body and Hand Pose Estimation Examples.}
Based on the inpainted images in Fig.~\ref{fig:inpainting}, we estimate the initial 3D SMPL-X body pose together with articulated hand poses.
The estimated poses capture coarse body configuration and hand-object alignment in image space, including which hand is used and its approximate contact location.
These estimates serve as strong initialization for our geometry-aware body refinement, significantly improving optimization stability, accelerating convergence, and reducing failure cases such as incorrect hand assignment or implausible body configurations.

\begin{figure}[!t]
    \centering
    \includegraphics[width=\linewidth]{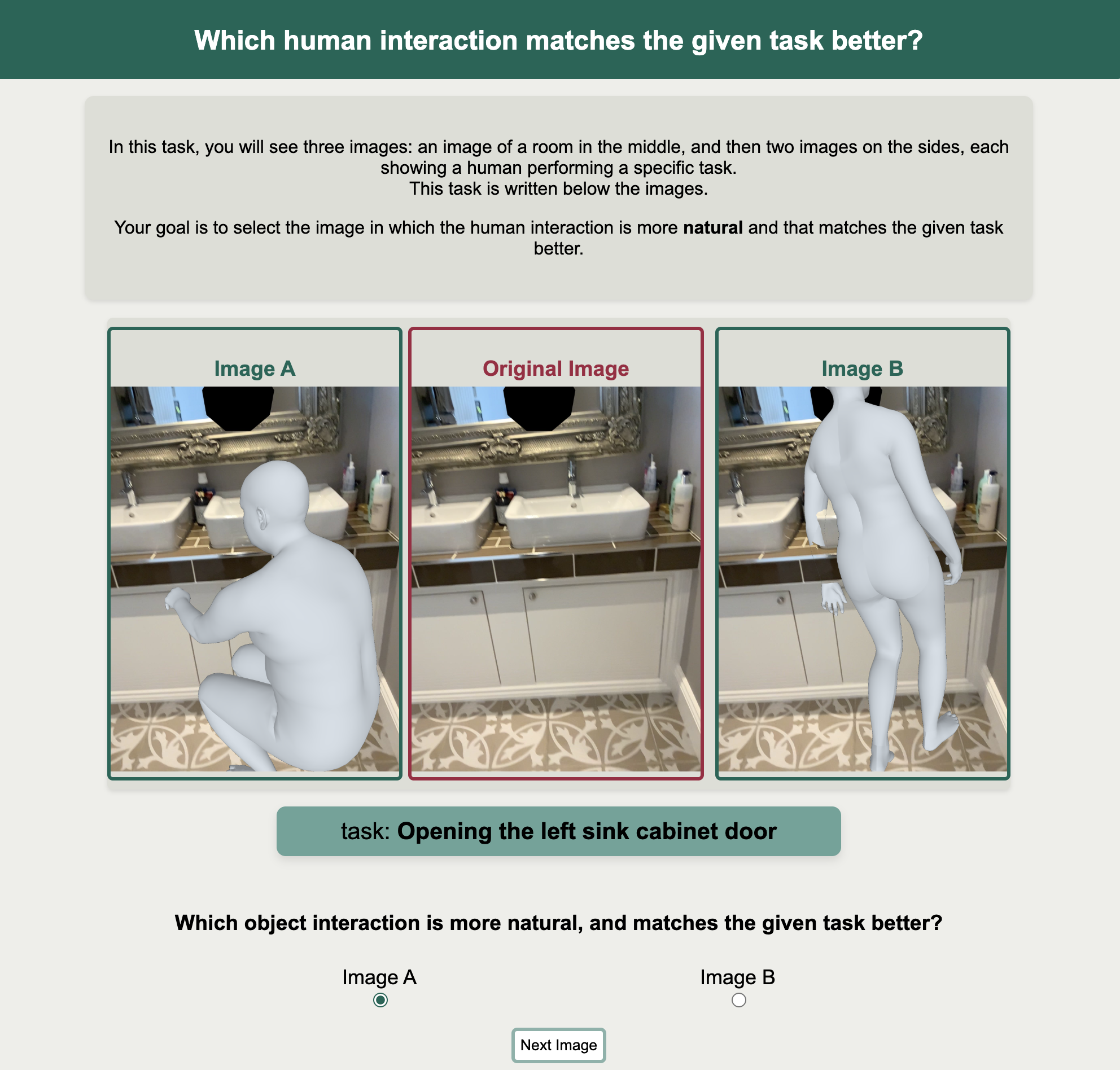}
    \caption{
        Layout of the perceptual study. Below the instructions, participants are presented with a target task label and three images: the original empty scene in the middle, and two candidate images on the sides depicting rendered human-scene interactions.
    }
    \label{fig:mturk_layout}
\end{figure}

\section{User Study Details}
\label{appendix:mturk}

We conduct a perceptual study on the Amazon Mechanical Turk platform over results rendered in 30 different scenes, evaluating a functional and a non-functional interaction prompt for each scene.
During the study, we present users with paired results—one from our method and one from a baseline. Users choose the result they prefer according to our criteria, and we report the percentage of cases in which the baseline is preferred over our method.
The layout of the perceptual study is shown in Fig.~\ref{fig:mturk_layout}.

We take several precautions in our study design to ensure reliable results. We only allow participants that are experienced ($\ge 5000$ accepted submissions) and highly rated ($\ge 98\%$ acceptance rate).
Each assignment contains 36 comparisons, i.e. pairs of images. The first three are intended as warm-up tasks, and the answers to these are discarded during evaluation. There are three so-called catch trials scattered among the remainder of the assignment. These are intentionally very obvious comparisons that help us identify participants who are providing random inputs. We discard all submissions where even a single one of the three catch trials is failed: 25 out of a total of 120 completions. To further reduce bias, the order of the comparisons is shuffled within an assignment, and the two sides of each comparison are randomly swapped too.

\end{document}